\documentclass[conference,a4paper]{IEEEtran}
\IEEEoverridecommandlockouts

\usepackage{cite}
\usepackage{amsmath,amssymb,amsfonts}
\usepackage{dsfont}
\usepackage{algorithm}
\usepackage{algpseudocode}
\usepackage{graphicx}
\usepackage{textcomp}
\usepackage{booktabs}
\usepackage{array}
\usepackage{xcolor}
\usepackage{tikz}
\usetikzlibrary{positioning,arrows.meta,fit,backgrounds}
\usepackage[hidelinks]{hyperref}
\usepackage{enumitem}

\newif\iftightpars \tightparstrue
\AddToHook{para/begin}{\iftightpars\looseness=-1\fi}
\AddToHook{env/figure/begin}{\tightparsfalse}
\AddToHook{env/figure*/begin}{\tightparsfalse}
\AddToHook{env/table/begin}{\tightparsfalse}
\AddToHook{env/table*/begin}{\tightparsfalse}

\begin{document}
\bstctlcite{IEEEexample:BSTcontrol}

\title{CanopyFormer: Attention-Guided SegFormer for Explainable Forest Cover Segmentation}

\author{
\IEEEauthorblockN{%
Dhinanjaya Fernando,
Dinura Ginige,
Kalana Lakshan,
Chanupa Gurusinghe,
Lasana Pahanga,\\
Dileesha Manamperi,
Thimira Sahan,
Sandareka Wickramanayake}
\IEEEauthorblockA{Dept.\ of Computer Science \& Engineering, University of Moratuwa, Moratuwa, Sri Lanka\\
\{dhinanjayaf.23, dinurag.23, kalanal.23, chanupag.23, lasanap.23,\\
dileesha.26, thimira.22, sandarekaw\}@cse.mrt.ac.lk
}
}

\maketitle

\begin{abstract}
In transformer segmentation models, attention is usually visualized after training, not supervised. Attention supervision exists for natural images and driving, but aerial forest segmentation uses attention only architecturally and judges interpretability qualitatively. We present CanopyFormer, an attention-guided SegFormer-B0 whose Attention Consistency Loss aligns its attention with the ground-truth forest mask, so its attention concentrates on canopy. Because SegFormer uses spatial-reduction attention, we adapt Gradient-weighted Attention Rollout by restricting it to the final encoder stage, where attention is square. We introduce Average Attention-Mask Overlap (AAMO) to measure alignment and add a deletion test of faithfulness. On 5{,}108 images, the method raises AAMO from 0.0334 to 0.7476, about 22 times that of a vanilla SegFormer-B0, at a modest Dice cost (from 0.8743 to 0.8577). The deletion test shows the gain is in alignment, not faithfulness: removing the most-attended quarter of the patches lowers the forest score by 12.1 percent, less than removing random patches (33.0 percent). Supervision adds no inference cost, but double backpropagation makes a training step 2.6 to 2.8 times slower. The alignment gain holds on a second dataset, LoveDA. A Boundary Refinement Module gives no consistent Dice gain and lowers AAMO, and mean squared error clearly outperforms Kullback-Leibler divergence as the attention loss.
\end{abstract}


\section{Introduction}

Forest-cover segmentation from aerial and satellite imagery supports ecological monitoring, deforestation tracking, and land management, and up-to-date land-cover and land-use maps underpin regional planning~\cite{meedeniya2020landuse}. Convolutional neural networks (CNNs) such as U-Net~\cite{ronneberger2015unet} long dominated the task, and transformers such as SegFormer~\cite{xie2021segformer} are now adopted for their long-range context. However, most segmentation work, including recent forestry studies~\cite{foresanext2026,ppliteseg2026,attunet2022deforestation}, uses attention only to improve accuracy and judges explainability, if at all, by inspecting heatmaps after training. A review confirms this: over 90\% of remote-sensing explainable artificial intelligence (XAI) studies evaluate explanations only anecdotally~\cite{hohl2024xairs}.

Because attention is not guided during training, it can drift to task-irrelevant regions such as image corners (Fig.~\ref{fig:qual}) instead of forest, which undermines trust in the model's explanations. Existing studies do not quantitatively evaluate the alignment or faithfulness of attention in forest segmentation.

Mask alignment is how well the thresholded attention map overlaps the ground-truth forest mask. It measures how plausible the explanation is, and Average Attention-Mask Overlap (AAMO) quantifies it (Section~\ref{sec:aamo}). Attention faithfulness is how well the map reflects the regions the prediction actually depends on~\cite{jacovi2020faithful}: removing highly attended pixels should change the prediction more than removing others. Alignment does not imply faithfulness~\cite{jain2019attention}, so we test faithfulness separately with a deletion test (Section~\ref{sec:deletion}).

We use an XAI method inside the training objective, not only for visualization. Gradient-weighted Attention Rollout~\cite{chefer2021transformer,abnar2020rollout} extracts an attention map from SegFormer's encoder, and an Attention Consistency Loss compares it with a Gaussian-smoothed ground-truth mask, alongside the segmentation loss. Because the map is itself defined through a gradient, training requires second-order gradients. We call the resulting model CanopyFormer.

This paper makes four contributions: (1) an Attention Consistency Loss that supervises SegFormer-B0's attention with the ground-truth forest mask during training, bringing attention supervision to the forestry domain; (2) an adaptation of Gradient-weighted Attention Rollout to SegFormer's spatial-reduction attention (SRA), restricted to the only encoder stage where rollout is well defined, with every implementation choice specified; (3) AAMO, a mask-alignment metric reported alongside segmentation metrics and paired with a deletion-based faithfulness test; and (4) an empirical study on 5{,}108 images based on the DeepGlobe 2018 dataset~\cite{DeepGlobe18}, with baselines, loss-weight sweeps, mean squared error (MSE) versus Kullback-Leibler (KL) divergence, a three-seed ablation repeated on LoveDA, and overhead and overfitting checks.\footnote{Code (anonymized for double-blind review): \url{https://anonymous.4open.science/r/explainable-segformer-forest-cover-C734/}}

\section{Related Work}
\label{sec:related}

Forest-cover segmentation has been treated almost only as an accuracy problem: better encoders, better attention modules, and larger datasets, with U-Net~\cite{ronneberger2015unet} as the standard baseline. Related land-cover studies in Sri Lanka, for example, segment rubber plantations with VGG-16 + FCN-8 and U-Net~\cite{mahakalanda2022rubber} and paddy fields with Mask R-CNN~\cite{meedeniya2022paddy}, judged by accuracy alone. What the network attends to, and why, is rarely examined: in remote sensing, explanations are mostly post-hoc and anecdotal~\cite{hohl2024xairs}. The closest work falls into two families.

\subsection{Attention as Architecture}
The first family uses attention to refine features and judges it only by segmentation accuracy. ForResANeXt~\cite{foresanext2026} adds lightweight attention and attention-gated skip connections to residual blocks for satellite forest segmentation. Attention-Refined PP-LiteSeg~\cite{ppliteseg2026} fuses channel, spatial, and pixel attention for cross-regional unmanned aerial vehicle (UAV) forest segmentation. An attention U-Net~\cite{attunet2022deforestation} adds attention gates for Sentinel-2 deforestation mapping. In all three, attention is neither supervised nor compared with any ground truth: it is an architectural convenience, not an explanation.

\subsection{Attention as Explainability}
The second family supervises attention as an interpretability output, but outside forest segmentation and without an attention-mask overlap metric. GAIN~\cite{li2018gain}, the closest method, trains a CNN so that its Grad-CAM-style attention covers the whole object, optionally with pixel masks, for weakly supervised segmentation; it reports only mean intersection over union (mIoU) and does not address transformers. Porres et al.~\cite{guiding2024driving} align the attention of driving models with salient semantic maps, but evaluate driving, not segmentation.

\subsection{Attention Extraction and Evaluation}
Attention Rollout~\cite{abnar2020rollout} multiplies attention matrices across layers to approximate token influence; its gradient-weighted variant~\cite{chefer2021transformer}, in the spirit of Grad-CAM~\cite{selvaraju2017gradcam}, first weights the heads by the target gradient. Both assume square, token-consistent self-attention in every layer, which SegFormer's Efficient Self-Attention violates (Section~\ref{sec:sra}). Attention maps are usually judged visually; numeric evaluation, when done, tests faithfulness by deleting the most relevant pixels~\cite{samek2017evaluating} rather than measuring overlap with a ground truth. AAMO (Section~\ref{sec:aamo}) measures that overlap, and we also run a deletion test (Section~\ref{sec:deletion}).

\subsection{Comparison and Research Gap}
\label{sec:comparison}
Table~\ref{tab:compare} compares our method with the closest work along five axes. No prior work combines (a) attention supervision during training, (b) forest segmentation, and (c) a quantitative alignment metric reported with segmentation metrics; ours does, and adds a faithfulness test. The same review confirms (a) and (c) field-wide: in remote sensing, only a few studies train explanations jointly with the model, and only 15 evaluate them quantitatively~\cite{hohl2024xairs}.

\begin{table*}[t]
\caption{Comparison with the closest land-cover, attention-supervision, and explainability work.}
\label{tab:compare}
\centering
\renewcommand{\arraystretch}{1.0}
\setlength{\tabcolsep}{3pt}
\begin{tabular}{|l|l|l|l|l|l|}
\hline
\textbf{Method} & \textbf{Attention target} & \textbf{Architecture} & \textbf{Domain / task} & \textbf{Explanation method} & \textbf{Evaluation metric} \\
\hline
ForResANeXt~\cite{foresanext2026} & None (architectural) & CNN + attention & Satellite forest seg. & None & Segmentation metrics \\
\hline
PP-LiteSeg~\cite{ppliteseg2026} & None (architectural) & CNN + attention & UAV forest seg. & None & Segmentation metrics \\
\hline
Attention U-Net~\cite{attunet2022deforestation} & None (architectural) & CNN + attention gates & Sentinel-2 deforestation & None & Segmentation metrics \\
\hline
Mahakalanda et al.~\cite{mahakalanda2022rubber} & None & CNN (FCN-8, U-Net) & Sentinel-2 rubber & None & Accuracy, IoU, Dice \\
\hline
Meedeniya et al.~\cite{meedeniya2022paddy} & None & Mask R-CNN & Sentinel-2 paddy & None & Average precision \\
\hline
Attention Rollout~\cite{abnar2020rollout} & None (post-hoc) & Transformer & Natural language & Rollout / flow & Agreement with ablation \\
\hline
Chefer et al.~\cite{chefer2021transformer} & None (post-hoc) & ViT & Image classification & Relevance propagation & Perturbation tests \\
\hline
GAIN~\cite{li2018gain} & Self-guidance, masks & CNN & Weakly supervised seg. & Grad-CAM-style & Segmentation mIoU \\
\hline
Porres et al.~\cite{guiding2024driving} & Salient semantic maps & CNN + transformer & Driving (CARLA) & Attention weights & Driving benchmarks \\
\hline
\textbf{Ours} & \textbf{Smoothed forest mask} & \textbf{SegFormer-B0 (SRA)} & \textbf{Aerial forest seg.} & \textbf{Stage-4 rollout} & \textbf{Dice, AAMO, deletion} \\
\hline
\end{tabular}
\end{table*}

\section{Method}
\label{sec:method}
\begin{figure}[t]
\centering
\input{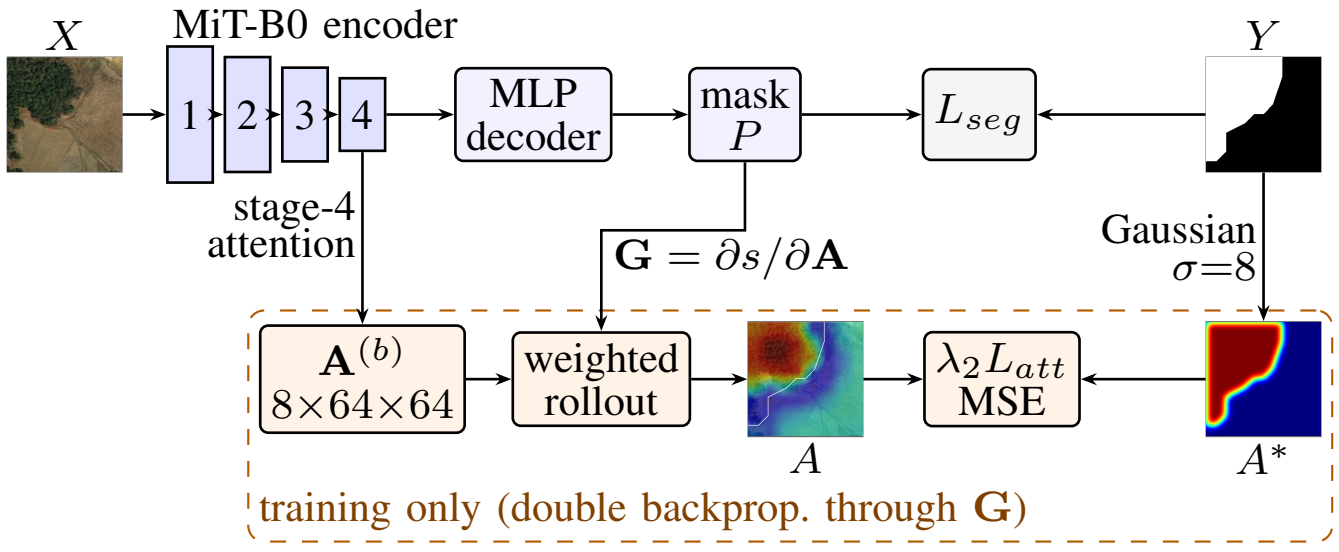}
\caption{Method overview: inference path (top) and training-only path (bottom).}
\label{fig:pipeline}
\end{figure}

The pipeline (Fig.~\ref{fig:pipeline}) adds a training-only attention-supervision path to SegFormer-B0, so inference is a standard SegFormer-B0 with no added cost. The MiT-B0 encoder and a multilayer perceptron (MLP) decoder map a red-green-blue (RGB) image $X$ to the predicted mask $P$. In parallel, an adapted Gradient-weighted Attention Rollout turns the stage-4 attention into a map $A$, and the Attention Consistency Loss $L_{att}$, added to the segmentation loss, pulls $A$ towards a Gaussian-smoothed ground-truth mask $A^{*}$. In Fig.~\ref{fig:pipeline}, the bottom row runs only in training; thumbnails show a test tile.

\subsection{Stage-4 Gradient-Weighted Attention Rollout}
\label{sec:sra}\label{sec:rollout}
Rollout's recursive matrix product needs square, token-consistent attention. SegFormer's Efficient Self-Attention shrinks the keys and values by a spatial-reduction ratio ($\mathrm{sr}=8,4,2$ in stages 1--3), so those attention matrices are rectangular. Only stage 4 ($\mathrm{sr}=1$) has true self-attention: for a $256\times256$ input, $n=64$ tokens on an $8\times8$ grid, with $H=8$ heads in each of its two blocks $b\in\{1,2\}$. We verified these shapes empirically and therefore restricted rollout to stage 4; Section~\ref{sec:rollout-diag} estimates the signal this leaves out. Forward hooks capture each block's softmax attention $\mathbf{A}^{(b)}\in[0,1]^{H\times n\times n}$, and the map is built in four steps.

(i) We explain the total predicted forest probability
\begin{equation}
s=\sum_{u}\operatorname{softmax}(\mathbf{z}_u)_{\mathrm{forest}},
\end{equation}
where $u$ ranges over the $64\times64$ decoder grid, $\mathbf{z}_u$ are the two-class decoder logits at $u$, and the subscript takes the forest-class probability, so one backward pass explains the whole forest prediction rather than one pixel.

(ii) With $\mathbf{G}^{(b)}=\partial s/\partial\mathbf{A}^{(b)}$, each attention entry is weighted by its own gradient and the heads are averaged~\cite{chefer2021transformer}:
\begin{equation}
\bar{A}^{(b)}=\frac{1}{H}\sum_{h=1}^{H}\big(\mathbf{G}^{(b)}_h\odot\mathbf{A}^{(b)}_h\big)^{+}\in\mathbb{R}^{n\times n}.
\end{equation}
where $h$ indexes the $H$ heads, $\odot$ is the element-wise product, and $(\cdot)^{+}$ zeroes negative entries. Unlike Grad-CAM, the gradient is not pooled: each query-key entry keeps its weight.

(iii) Each block adds the identity for its residual connection and is row-normalized, $\tilde{A}^{(b)}=D_b^{-1}(\bar{A}^{(b)}+I)$ with $D_b$ the diagonal matrix of row sums, and the two blocks are multiplied, $R=\tilde{A}^{(2)}\tilde{A}^{(1)}$.

\begin{figure}[t]
\centering
\input{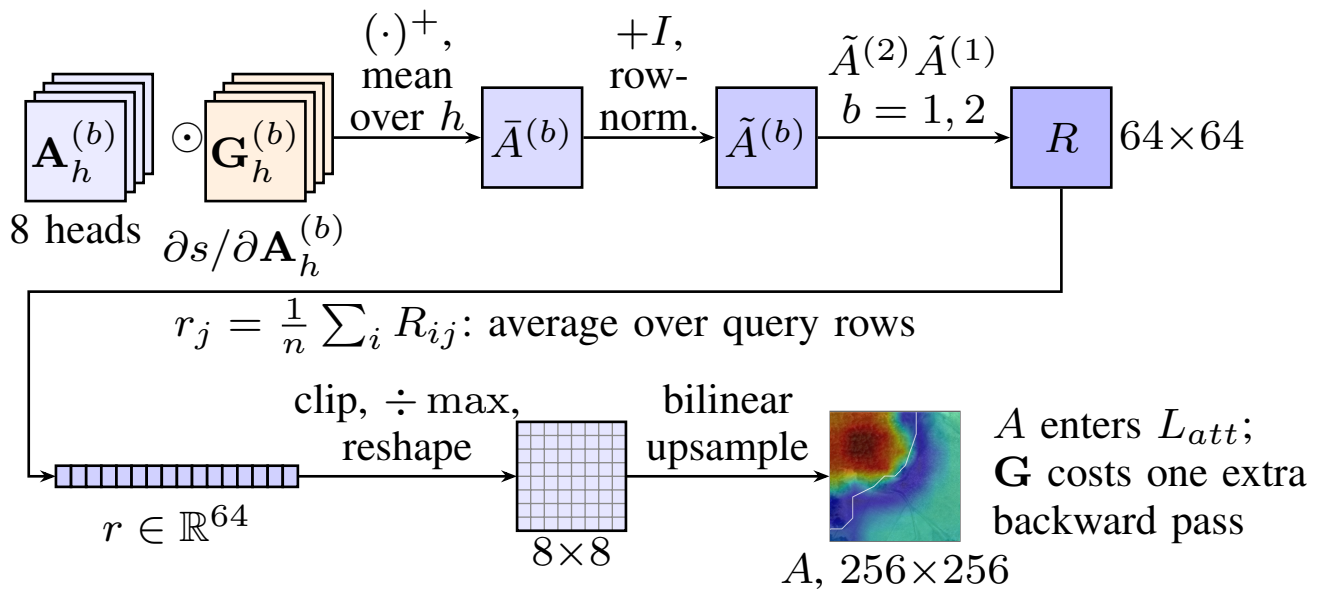}
\caption{Gradient-weighted attention rollout on SegFormer stage 4.}
\label{fig:rollout}
\end{figure}

(iv) SegFormer has no class token, so we average $R$ over all query rows, $r_j=\frac{1}{n}\sum_{i}R_{ij}$. The vector $r$ is clipped at zero, divided by its maximum, reshaped to $8\times8$, and bilinearly upsampled to $A\in[0,1]^{256\times256}$. Fig.~\ref{fig:rollout} shows steps (ii) to (iv).

\subsection{Loss Terms and Double Backpropagation}
\label{sec:latt}\label{sec:dbp}
The soft target $A^{*}=\mathrm{Gaussian}(Y,\sigma)$ blurs the ground-truth mask $Y\in\{0,1\}^{256\times256}$ with a separable kernel of radius $3\sigma$ and divides by its maximum, so $A^{*}\in[0,1]$; $\sigma=8$ was tuned on the validation set. We implement two forms: a mean squared error averaged over pixels,
\begin{equation}
L_{att}^{\mathrm{MSE}} = \mathrm{MSE}(A, A^{*}),
\end{equation}
where $A$ is the rollout map and $A^{*}$ the soft target, and a KL divergence that treats both as distributions over pixels,
\begin{equation}
L_{att}^{\mathrm{KL}} = \mathrm{KL}(A \,\Vert\, A^{*}).
\end{equation}
MSE is clearly better (Section~\ref{sec:msekl}), so $L_{att}$ denotes $L_{att}^{\mathrm{MSE}}$ unless stated otherwise.

Most errors occur near canopy edges, so, following edge-focused losses~\cite{kervadec2019boundary}, the Boundary Refinement Module adds a Dice term on the boundaries of prediction and ground truth:
\begin{equation}
L_{boundary} = 1 - \frac{2\,|\partial P \cap \partial Y|}{|\partial P| + |\partial Y|},
\end{equation}
where $P$ is the predicted forest-probability map, $Y$ the ground-truth mask, $\partial(\cdot)$ the morphological gradient ($3\times3$ kernel), which keeps only boundary pixels, $\cap$ the pixel-wise product, and $|\cdot|$ the sum over pixels; tables abbreviate the loss as $L_{bnd}$. It is added on top of $L_{att}$, not instead of it, with weight $\lambda_3$ chosen on the validation set (Section~\ref{sec:l3sweep}).

With a Dice loss $L_{dice}$ and binary cross-entropy $L_{bce}$ on $P$ and $Y$, the full objective is
\begin{equation}
L = L_{dice} + \lambda_1 L_{bce} + \lambda_2 L_{att} + \lambda_3 L_{boundary},
\end{equation}
where $\lambda_1$, $\lambda_2$ and $\lambda_3$ weight the terms, with $\lambda_1 = 1$. Because $A$ is itself built from a gradient, backpropagating $L_{att}$ needs a second derivative. We compute $\mathbf{G}^{(b)}$ with \texttt{create\_graph=True}, so $A$ stays in the autograd graph and one \texttt{backward()} reaches every parameter. This makes a training step 2.6--2.8$\times$ slower (Section~\ref{sec:overhead}).

\section{Average Attention-Mask Overlap}
\label{sec:aamo}
AAMO makes attention-mask alignment reportable and comparable. Given an attention map min-max rescaled to $A\in[0,1]^{256\times256}$ and the mask $Y$, we first threshold the map at $\tau=0.5$,
\begin{equation}
A_{\mathrm{thr}} = \mathds{1}[A \ge \tau],
\end{equation}
where $\mathds{1}[\cdot]$ is 1 at pixels where the condition holds and 0 elsewhere, and then measure how much of the mask $Y$ the thresholded map covers:
\begin{equation}
\mathrm{AAMO} = \frac{|A_{\mathrm{thr}} \cap Y|}{|Y|}.
\end{equation}
where $\cap$ keeps the pixels set in both maps and $|\cdot|$ counts pixels. AAMO is the recall of the forest mask under thresholded attention: $1.0$ means full coverage and $0.0$ no overlap. Images with an empty mask contribute 0, and we report the mean over all images of a split.

As a recall, AAMO does not penalize attention that spills onto nonforest areas. In practice, 81\% of the thresholded CanopyFormer map lay on forest, against 61\% for a map marking every pixel and under 60\% for the vanilla map, and the ranking held for $\tau$ from 0.3 to 0.7. More importantly, it measures plausibility, not faithfulness: a map can overlap the mask while the prediction relies on other evidence~\cite{jain2019attention}. Since $L_{att}$ directly optimizes overlap with $A^{*}$, part of the AAMO gain is expected by construction. We therefore pair AAMO with a deletion test of faithfulness~\cite{samek2017evaluating} (Section~\ref{sec:deletion}).

\section{Experimental Setup}
\label{sec:setup}

We used 5{,}108 RGB aerial images ($256\times256$) with binary forest/nonforest masks, based on the DeepGlobe 2018 challenge~\cite{DeepGlobe18}.\footnote{Forest Aerial Images for Segmentation: \url{https://www.kaggle.com/datasets/quadeer15sh/augmented-forest-segmentation}.} They were split 70:15:15 (3576/766/766) with seed 42, so all models shared one test set; in the multi-seed runs, the seed also redrew the split (Section~\ref{sec:multiseed}). An audit script confirmed that all files were readable and all masks binary. Because the data are forest-majority (mean forest ratio $0.6117 \pm 0.3306$ per mask), we used Dice, intersection over union (IoU), and F1 score rather than pixel accuracy.

U-Net~\cite{ronneberger2015unet} and a vanilla SegFormer-B0 (no attention loss) isolated the effect of supervision, and DeepLabV3+~\cite{chen2018deeplabv3plus} with a MobileNetV3~\cite{howard2019mobilenetv3} backbone added a second CNN reference. SegFormer-B2~\cite{xie2021segformer} and Swin-T with UPerNet~\cite{liu2021swin,xiao2018upernet} served as larger transformer baselines, trained from ImageNet-pretrained encoders with the same split, epochs, optimizer, and checkpoint rule, at batch size 16 on a T4 graphics processing unit (GPU) (best epochs 8 and 12).

Besides Dice, IoU, and F1, we report AAMO (alignment), the deletion score drop (faithfulness, Section~\ref{sec:deletion}), and parameters, GMACs (billions of multiply-accumulates), time, and peak memory (efficiency, Section~\ref{sec:overhead}).

SegFormer-B0 variants were trained for 20 epochs with AdamW (learning rate $6\times10^{-5}$), at batch size 16 for the vanilla model and 1 for the attention and boundary variants, whose rollout was computed per image. We used $\lambda_2 = 1.0$, chosen by the validation sweep (Section~\ref{sec:l2sweep}), and report each variant's best-validation checkpoint, since validation Dice peaked early (Section~\ref{sec:overfit}).

\section{Results}
\label{sec:results}

\subsection{Main Comparison}
\begin{figure}[t]
\centering
\includegraphics[width=0.9\columnwidth]{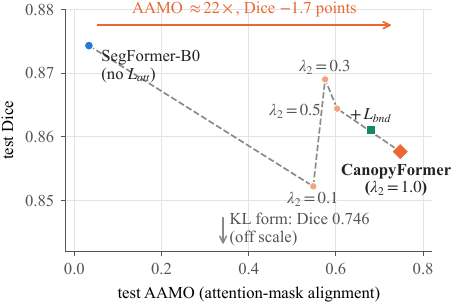}
\caption{Alignment against accuracy for the SegFormer-B0 variants.}
\label{fig:tradeoff}
\end{figure}

Table~\ref{tab:main} reports all models on the shared seed-42 test set. The Attention Consistency Loss raised AAMO about 22-fold while Dice stayed nearly flat. Fig.~\ref{fig:tradeoff} plots test Dice against test AAMO for every SegFormer-B0 variant at seed 42; the dashed path follows $\lambda_2$ from the vanilla model ($\lambda_2 = 0$) to CanopyFormer. The vanilla SegFormer-B0 already beat U-Net with under an eighth of its parameters. SegFormer-B2 and Swin-T + UPerNet were slightly more accurate, but 7 and 16 times larger and several times slower on a T4 GPU. SegFormer-B0's small size is what keeps per-image rollout and double backpropagation affordable. Against SegFormer-B2, CanopyFormer gave up about two Dice points for a seventh of the parameters and an attention map that follows the canopy.

\begin{table}[t]
\caption{Test results (seed 42, shared 766-image test set). The KL row changes only the form of $L_{att}$.}
\label{tab:main}
\centering
\renewcommand{\arraystretch}{1.0}
\setlength{\tabcolsep}{2pt}
\footnotesize
\begin{tabular*}{\columnwidth}{@{\extracolsep{\fill}}|l|c|c|c|c|c|}
\hline
\textbf{Model} & \textbf{Dice} & \textbf{IoU} & \textbf{F1} & \textbf{AAMO} & \textbf{Params} \\
\hline
U-Net (CNN) & 0.8615 & 0.7568 & 0.8615 & n/a & 31.0\,M \\
\hline
DeepLabV3+ (MNetV3) & 0.8716 & 0.7724 & 0.8716 & n/a & 11.0\,M \\
\hline
SegFormer-B2 & \textbf{0.8809} & \textbf{0.7872} & \textbf{0.8809} & n/a & 27.3\,M \\
\hline
Swin-T + UPerNet & 0.8793 & 0.7847 & 0.8793 & n/a & 58.9\,M \\
\hline
SegFormer-B0 & 0.8743 & 0.7766 & 0.8743 & 0.0334 & 3.7\,M \\
\hline
\;+ $L_{att}$ (MSE), CanopyFormer & 0.8577 & 0.7508 & 0.8577 & \textbf{0.7476} & 3.7\,M \\
\hline
\;+ $L_{att}$ (KL) & 0.7462 & 0.5951 & 0.7462 & 0.3412 & 3.7\,M \\
\hline
\;+ $L_{att}$ + $L_{bnd}$ ($\lambda_3{=}0.5$) & 0.8611 & 0.7560 & 0.8611 & 0.6805 & 3.7\,M \\
\hline
\end{tabular*}
\end{table}

\subsection{Loss-Weight Selection}
\label{sec:l2sweep}\label{sec:l3sweep}
Table~\ref{tab:weights} lists all weights tried. Raising $\lambda_2$ lifted AAMO while Dice barely moved, so we picked $\lambda_2 = 1.0$, the highest validation AAMO; validation and test agreed, so the test split played no part. With $\lambda_2$ fixed, we ranked $\lambda_3$ by validation Dice, then IoU. Seed 42 alone favored 0.2, which lost much alignment, but the three-seed mean favored 0.5, a weak preference of under one standard deviation. Weights of 0.2 or more also moved the best checkpoint much earlier.

\begin{table}[t]
\caption{Loss-weight selection (seed 42 unless noted); selected weights in bold.}
\label{tab:weights}
\centering
\renewcommand{\arraystretch}{1.0}
\setlength{\tabcolsep}{2pt}
\footnotesize
\begin{tabular*}{\columnwidth}{@{\extracolsep{\fill}}|l|c|c|c|c|c|c|}
\hline
 & \multicolumn{3}{c|}{\textbf{Validation}} & \multicolumn{2}{c|}{\textbf{Test}} & \textbf{Best} \\
\cline{2-4}\cline{5-6}
\textbf{Weight} & \textbf{Dice} & \textbf{AAMO} & \textbf{Dice (3 seeds)} & \textbf{Dice} & \textbf{AAMO} & \textbf{epoch} \\
\hline
$\lambda_2 = 0.1$ & 0.8664 & 0.5354 & -- & 0.8523 & 0.5481 & 4 \\
\hline
$\lambda_2 = 0.3$ & 0.8695 & 0.5801 & -- & 0.8690 & 0.5752 & 3 \\
\hline
$\lambda_2 = 0.5$ & 0.8661 & 0.6014 & -- & 0.8644 & 0.6028 & 4 \\
\hline
$\boldsymbol{\lambda_2 = 1.0}$ & 0.8626 & \textbf{0.7519} & -- & 0.8577 & 0.7476 & 11 \\
\hline
$\lambda_3 = 0.1$ & 0.8601 & -- & $0.8475 \pm 0.0123$ & 0.8558 & 0.7775 & 11 \\
\hline
$\lambda_3 = 0.2$ & \textbf{0.8717} & -- & $0.8512 \pm 0.0281$ & 0.8669 & 0.6218 & 4 \\
\hline
$\boldsymbol{\lambda_3 = 0.5}$ & 0.8685 & -- & $\mathbf{0.8582 \pm 0.0108}$ & 0.8611 & 0.6805 & 4 \\
\hline
\end{tabular*}
\end{table}

\subsection{Choice of Attention Loss}
\label{sec:msekl}
Replacing MSE with KL hurt every metric in Table~\ref{tab:main}: Dice fell by 11 points and AAMO by more than half. The gap dwarfed the seed-to-seed spread of the MSE model and already appeared on validation, so KL is simply a weaker objective, not one that generalizes worse. A likely reason is that KL penalizes the ratio $A/A^{*}$, which puts more gradient on low-target pixels where $A$ is nonzero, whereas MSE penalizes $|A-A^{*}|$ directly.

\subsection{Multi-Seed Ablation and Cross-Dataset Check}
\label{sec:multiseed}\label{sec:loveda}
Table~\ref{tab:multiseed} summarizes seeds 42, 43 and 44. On DeepGlobe, the seed also redrew the data split. The vanilla and attention models were stable across seeds. The boundary model was not: at $\lambda_3 = 0.2$, seed 43 fell about 4 Dice points below the other two, while $\lambda_3 = 0.5$, the setting in Table~\ref{tab:multiseed}, varied less. AAMO was noisier, and the two supervised models swapped order at seed 43.

We also repeated the three configurations, with unchanged code and settings, on the LoveDA land-cover dataset~\cite{wang2021loveda},\footnote{LoveDA: 0.3\,m RGB tiles of $1024\times1024$ pixels from Nanjing, Changzhou and Wuhan, China, with seven classes; we cut $256\times256$ patches with 5--95\% forest. Data: \url{https://zenodo.org/records/5706578}.} with forest as the positive class, splits of the same sizes, and the LoveDA validation tiles as the test set. Accuracy was lower, but the pattern held: $L_{att}$ raised AAMO about 28-fold in every seed at a small Dice cost, while $L_{bnd}$ lowered both Dice and AAMO. Reruns of two seeds on other GPUs shifted test Dice by up to 5.5 points without changing either conclusion.

\begin{table}[t]
\caption{Mean $\pm$ standard deviation over three seeds.}
\label{tab:multiseed}
\centering
\renewcommand{\arraystretch}{1.0}
\setlength{\tabcolsep}{2pt}
\footnotesize
\begin{tabular*}{\columnwidth}{@{\extracolsep{\fill}}|l|c|c|c|}
\hline
\textbf{Config} & \textbf{Dice} & \textbf{IoU} & \textbf{AAMO} \\
\hline
\multicolumn{4}{|c|}{\textbf{DeepGlobe-based (seed also redraws the split)}} \\
\hline
Vanilla & $0.8723 \pm 0.0024$ & $0.7735 \pm 0.0038$ & $0.0435 \pm 0.0167$ \\
\hline
$+\,L_{att}$ (MSE) & $0.8564 \pm 0.0026$ & $0.7488 \pm 0.0039$ & $0.7028 \pm 0.0392$ \\
\hline
$+\,L_{att}+L_{bnd}$ & $0.8505 \pm 0.0139$ & $0.7399 \pm 0.0208$ & $0.6881 \pm 0.0719$ \\
\hline
\multicolumn{4}{|c|}{\textbf{LoveDA (one fixed split)}} \\
\hline
Vanilla & $0.7094 \pm 0.0077$ & $0.5497 \pm 0.0092$ & $0.0217 \pm 0.0019$ \\
\hline
$+\,L_{att}$ (MSE) & $0.6955 \pm 0.0012$ & $0.5332 \pm 0.0014$ & $0.6209 \pm 0.0520$ \\
\hline
$+\,L_{att}+L_{bnd}$ & $0.6706 \pm 0.0220$ & $0.5048 \pm 0.0251$ & $0.5924 \pm 0.0134$ \\
\hline
\end{tabular*}
\end{table}

\subsection{Faithfulness: Deletion Test}
\label{sec:deletion}
We ran a deletion test~\cite{samek2017evaluating} on the 766 test images with the seed-42 vanilla and attention models. Each image was split into the $8\times8$ grid of $32\times32$ patches that matches the stage-4 tokens; we replaced the top-$k$\% of patches ($k = 10, 25, 50$, i.e., 6, 16 and 32 of the 64 patches) with the mean color or a blurred copy (Gaussian, $\sigma=16$ pixels) and recorded the relative drop of $s$. Patches were ranked by mean attention, at random, or by ground-truth forest fraction (forest-first). Forest-first is the key control: since $L_{att}$ pulls attention onto forest, attention-ranked deletion must beat it to show that the map ranks patches by reliance, not by class.

\begin{figure}[t]
\centering
\includegraphics[width=0.9\columnwidth]{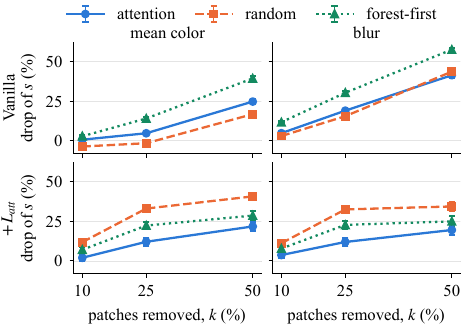}
\caption{Deletion test: drop of the forest score $s$ for each deletion order.}
\label{fig:deletion}
\end{figure}

Fig.~\ref{fig:deletion} plots the mean relative drop of $s$ ($\pm1$ standard error), one row per model and one column per replacement. The result is clearly negative. For the supervised model, deleting the most-attended patches was the weakest order at every budget and with both replacements: removing the top 25\% lowered $s$ by only 12.1\%, against 33.0\% for random patches, and forest-first sat in between. For the vanilla model, attention stayed close to random and well below forest-first. Neither map ranks patches by how much the prediction depends on them: the 22-fold AAMO gain is a gain in alignment, not faithfulness. One untested explanation is redundancy: the supervised map covers canopy interiors, where neighboring patches support the same prediction. The pipeline reproduced Table~\ref{tab:main}'s AAMO exactly.

\subsection{Computational Overhead}
\label{sec:overhead}
We also timed one batch-1 training step of each model on a laptop central processing unit (CPU; 12th-generation Intel Core, 4 threads) and a T4 GPU (Table~\ref{tab:overhead}). The cost lay in training: a step took 2.6--2.8$\times$ as long for only a third more compute, because the second backward pass consists of many small operations. An explanation map cost one extra backward pass; the boundary module added 4\% step time.

\begin{table}[t]
\caption{Computational overhead of attention supervision.}
\label{tab:overhead}
\centering
\renewcommand{\arraystretch}{1.0}
\setlength{\tabcolsep}{2pt}
\footnotesize
\begin{tabular*}{\columnwidth}{@{\extracolsep{\fill}}|l|c|c|c|c|c|c|}
\hline
 & \multicolumn{3}{c|}{\textbf{CPU (laptop)}} & \multicolumn{3}{c|}{\textbf{GPU}} \\
\cline{2-4}\cline{5-7}
\textbf{Quantity} & \textbf{Vanilla} & \textbf{$+\,L_{att}$} & \textbf{Ratio} & \textbf{Vanilla} & \textbf{$+\,L_{att}$} & \textbf{Ratio} \\
\hline
Inference (ms/image) & 72 & 72 & 1.00 & 9.0--11.8 & 9.7 & $\approx$1 \\
\hline
Training step (ms) & 401 & 1108 & 2.8 & 34 & 90 & 2.6 \\
\hline
Peak memory (MB) & 203$^{b}$ & 259$^{b}$ & 1.27 & 149 & 186 & 1.25 \\
\hline
Explanation map (ms) & 209 & 209 & -- & 39 & 39 & -- \\
\hline
Time per epoch (s)$^{a}$ & -- & -- & -- & 44 & 386 & 8.8 \\
\hline
Step compute (GMACs)$^{c}$ & 10.7 & 14.2 & 1.33 & -- & -- & -- \\
\hline
\end{tabular*}\\[2pt]
{\footnotesize\raggedright GPU batch-1 rows: T4. $^{a}$Full-scale runs, 3-seed mean (vanilla batch 16, attention batch 1). $^{b}$Growth over loaded model. $^{c}$Device-independent.\par}
\end{table}

\subsection{Diagnostics and Qualitative Results}
\label{sec:overfit}\label{sec:rollout-diag}
Every run overfitted: validation loss turned upward after an early minimum while training metrics kept improving. The train/validation Dice gap widened most for the vanilla model, so $L_{att}$ and $L_{boundary}$ act as mild regularizers. In a single-seed grid, weight decay helped only the vanilla model; the attention model did best with a cosine schedule.
To check what the stage-4 restriction discards, we averaged each stage's raw attention onto the $8\times8$ grid and correlated it with the downsampled masks. Stages 1--3 correlated only weakly ($r \le 0.19$) and stage 4 strongly ($r = 0.52$), so the restriction mainly removes mask-irrelevant signal.

\begin{figure*}[t]
\centering
\input{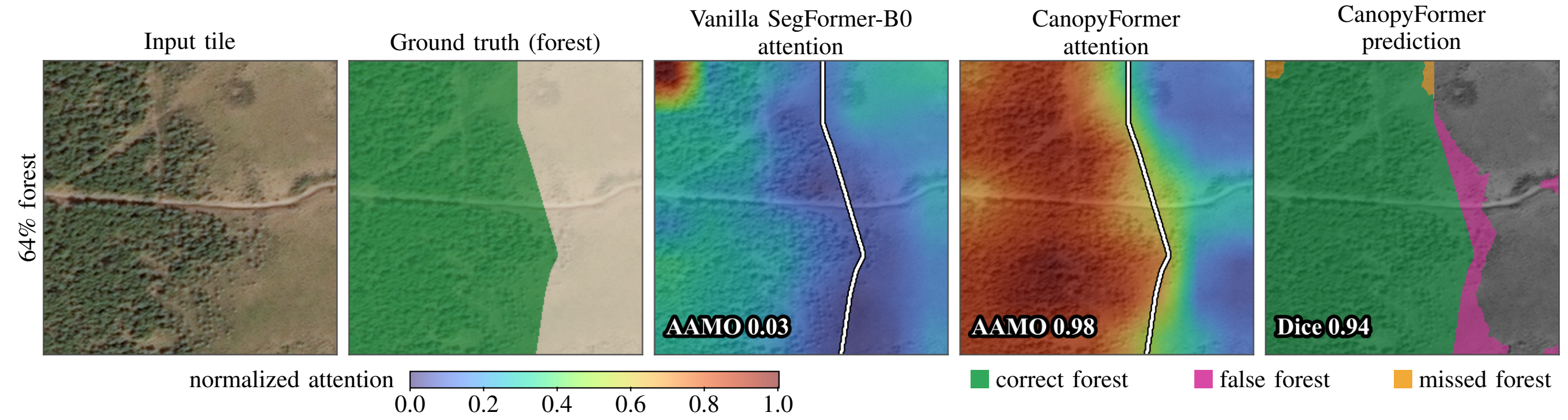}
\caption{Vanilla and CanopyFormer attention on a test tile.}
\label{fig:qual}
\end{figure*}

Fig.~\ref{fig:qual} shows a test tile with 30--70\% forest, its mask (white line), both attention maps scaled to $[0,1]$, and CanopyFormer's prediction, all from the seed-43 models. Per-image AAMO rises from 0.03 to 0.98. CanopyFormer's map covers the canopy and falls off at the forest boundary. The vanilla attention collects in a corner hot spot rather than on roads or shadows, a pattern that persisted at full scale and so is unlikely to come from undertraining.

\section{Discussion}
\label{sec:discussion}

The large AAMO gain at a small Dice cost, and the steady AAMO rise over a nearly flat Dice range in the $\lambda_2$ sweep, suggest a tunable trade-off between alignment and accuracy. By the deletion test, the alignment gained by $L_{att}$ is not faithfulness, so AAMO does not show the model's evidence.

Adding $L_{boundary}$ did not raise Dice reliably, lowered AAMO, and moved the best checkpoint much earlier. A possible cause is gradient competition in the shared encoder: sharp edge features favored by $L_{boundary}$ may conflict with the canopy-focused target of $L_{att}$.

The deployed model is a plain SegFormer-B0, small enough for a forest-change web tool, on-board drone use, or flagging tiles where explanation and mask disagree. Before real deforestation decisions, it needs validation on the target region and sensor (e.g., Sri Lankan forests) and human review.

Several limitations remain. The transformer baselines were single-seed, and DeepLabV3+ varied only its initialization. The boundary model's variance came mainly from seed 43, whose best checkpoint arrived much later and whose test Dice was lower, although the same split trained normally without $L_{boundary}$. AAMO is partly optimized by construction, and the single-seed, patch-level deletion test could rank the maps differently under another measure.

\section{Conclusion and Future Work}
\label{sec:conclusion}
We supervised SegFormer-B0's attention with the forest mask through an Attention Consistency Loss, adapted Gradient-weighted Attention Rollout to spatial-reduction attention, and introduced AAMO. The loss raised AAMO about 22-fold at a small Dice cost, also on LoveDA; it slows training but not inference. A deletion test showed this to be a gain in alignment, not faithfulness. The Boundary Refinement Module gave no consistent Dice gain and reduced AAMO, and MSE beat KL as the attention loss. Future work will train for faithfulness, decouple the boundary head, add seeds and baselines, and validate on Sri Lankan multispectral imagery.

\tightparsfalse 
\bibliographystyle{IEEEtran}
\bibliography{references}

\end{document}